# Variable-aware representation, not relation typing, explains the performance of an event-graph model for ICU mortality prediction: temporal and external validation

Article type: Original Article

Mohammad Nasir Uddin[1]*, Eklachur Rahman Bhuiyan[2], Rahnuma Tabassum Orpita[3]

[1] College of Business, DBA-BIDA Program, Westcliff University, Los Angeles, CA 90035, USA. [2] School of Information Technology, Washington University of Science and Technology, Alexandria, VA 22314, USA. [3] Department of Computer Science and Engineering, Northern University Bangladesh, Dhaka 1230, Bangladesh.

*Corresponding author: Mohammad Nasir Uddin (m.uddin.258@westcliff.edu)

ORCID: Mohammad Nasir Uddin, 0009-0009-0990-4616; Eklachur Rahman Bhuiyan, 0009-0006-1424-0373; Rahnuma Tabassum Orpita, 0009-0008-8399-6945

**Acknowledgements:** The authors thank the PhysioNet team, the MIT Laboratory for Computational Physiology and the eICU Research Institute for maintaining MIMIC-IV and eICU-CRD.

## Abstract

Event-graph neural networks represent intensive care unit (ICU) records as typed entities linked by timestamped observations, and their gains are often attributed to relation-specific message passing. We tested which components matter once leakage is removed and validation is temporal and external. In 31,142 first ICU stays of at least 48 hours in the Medical Information Mart for Intensive Care IV (MIMIC-IV), models predicting post-48-hour in-hospital mortality were developed on 2008–2016 stays, tested on held-out 2008–2016 stays and on 2017–2019 and 2020–2022 stays, and applied without retraining to 54,696 stays from 205 hospitals in the eICU Collaborative Research Database. A leak in a previous version, total ICU length of stay used as an input, was removed. A variable-aware representation (a learned embedding of each node's variable plus per-variable normalisation) raised the mean single-model area under the receiver operating characteristic curve (AUROC) from 0.846 to 0.896 in 2020–2022 and from 0.699 to 0.809 externally. With it, relation-specific and shared (relation-collapsed) message passing differed by less than 0.005 AUROC in every test set, within a ±0.01 equivalence margin, although relation typing had appeared beneficial without it. The revised model's AUROC was 0.878–0.900 across MIMIC-IV periods, 0.15–0.19 above three severity scores and 0.007–0.011 below gradient boosting. Externally, the five-model average reached AUROC 0.825 versus 0.788 for the Acute Physiology and Chronic Health Evaluation (APACHE) IVa score on the 48,184 stays with an APACHE prediction. In this model, representation mattered more than message-passing architecture.

## 1 Introduction

Prediction of in-hospital mortality from intensive care unit (ICU) data is a well-studied task, and the electronic health records (EHRs) of ICU stays pose two structural difficulties for it. They are irregular: a heart rate may be charted every few minutes during resuscitation and every few hours overnight, while a laboratory value may be measured twice in a day or not at all (Shukla and Marlin 2018, 2021). And they are relational: a stay consists of a visit, monitored vital-sign channels, laboratory analytes and treatments connected by observations (Wanyan et al. 2021; Chen et al. 2024). Graph neural networks address both by representing a stay as a typed graph, and several recent ICU models report gains from adding graph or temporal-graph structure (Wanyan et al. 2021; Rocheteau et al. 2021; Chen et al. 2024; Li et al. 2024; Shi et al. 2024; Cifci et al. 2025; Rahman et al. 2026).

Such gains are easily overstated. Inputs whose availability at prediction time cannot be verified inflate performance (Kaufman et al. 2012; Kapoor and Narayanan 2023); baselines that are weaker than routinely achievable make any new architecture look better; and a single random split of one hospital's data says little about performance over time or in other hospitals. Architectural ablations add a subtler hazard: when a component is removed, the effect attributed to it may instead reflect what the remaining model can or cannot represent.

We previously reported an event-graph model, the continuous-time heterogeneous EHR graph (CT-HEG) with the CHIRP-Net graph attention network, in which every observation is an edge carrying its own timestamp and value, and concluded from capacity-controlled ablations that relation-specific (heterogeneous) message passing improved discrimination (Uddin et al. 2026). A subsequent audit found that the visit node of that model contained the total ICU length of stay, a quantity known only at discharge, and that its baselines were considerably weaker than properly constructed tabular models. Here we correct the model and re-evaluate it under a substantially stronger design: temporal validation across three periods of the Medical Information Mart for Intensive Care IV (MIMIC-IV), external validation in 205 hospitals of the eICU Collaborative Research Database (eICU-CRD), strong tabular, recurrent and clinical-score baselines, paired patient-level tests and calibration analysis.

The study makes four contributions. First, it documents two leakage audits, of input timing and of node features, and quantifies how the corrected model compares with strong baselines. Second, it identifies a variable-aware representation as the decisive change: in the previous architecture a vital-sign or laboratory node carried only six summary statistics, z-scored across all variables pooled, so the model could not tell a heart-rate node from an oxygen-saturation node except by value range. Third, it shows that the previously reported benefit of relation typing is no longer detectable once that representation is used, so that heterogeneous-versus-homogeneous ablations of this kind should be interpreted against the strongest available representation. Fourth, it shows that the corrected model's discrimination did not decrease across MIMIC-IV anchor-year groups and that, in pooled external validation, it discriminated better than the Acute Physiology and Chronic Health Evaluation (APACHE) IVa score, although its external calibration required correction. We do not claim that the model outperforms gradient boosting; it does not.

## 2 Related work

Sequential and irregular-time-series models. The gated recurrent unit with decay (GRU-D; Che et al. 2018) models missingness with a learned decay; the multi-time attention network (mTAND; Shukla and Marlin 2021) learns continuous-time attention over observed timestamps; ALNN (Bignoumba et al. 2024) aligns irregular variables; and MedGAITS (Wang et al. 2026) learns from incomplete records with a graph autoencoder. These represent the patient as a multivariate series rather than a typed graph.

EHR graph neural networks. HGM-CNN (Wanyan et al. 2021) embeds a heterogeneous graph of medical concepts; TRANS (Chen et al. 2024) models a record as a temporal heterogeneous graph of visit and event nodes; Time-aware HGT (Li et al. 2024) incorporates time into visit-level embeddings; and LSTM-GNN (Rocheteau et al. 2021) adds a patient-similarity graph. DyGraphTrans (Rahman et al. 2026), ICU-TGNN (Shi et al. 2024) and AGFN (Cifci et al. 2025) model temporal dynamics through snapshots, multitask transformers or adaptive similarity graphs. These models generally represent time at the level of visits, windows or snapshots; CT-HEG instead attaches each observation's timestamp to its own edge.

Tabular baselines and evaluation. Tree-based ensembles remain strong on tabular data (Grinsztajn et al. 2022), and summary-statistic gradient boosting is a demanding comparator for ICU prediction. Data leakage is a recognised source of irreproducible machine-learning results (Kapoor and Narayanan 2023), and clinical prediction models are expected to report calibration and temporal or external validation (Collins et al. 2024). Few EHR graph studies report all three of a strong tabular comparator, temporal validation and multi-hospital external validation; this study does.

# 3 Methods

## 3.1 Data sources, cohorts and outcome

Development data were MIMIC-IV v3.1 (Johnson et al. 2023, 2024), a de-identified database of admissions to Beth Israel Deaconess Medical Center, Boston, MA, USA. Following the 48-hour in-hospital-mortality convention of the MIMIC-III benchmark (Harutyunyan et al. 2019), we included adult patients (age ≥ 18 years), retained each patient's first ICU stay (earliest ICU admission time) and then required that stay to last at least 48 hours; patients whose first stay was shorter were excluded rather than replaced by a later stay. The cohort comprised 31,142 stays from 31,142 patients, with an in-hospital mortality of 13.4%.

External validation data were eICU-CRD v2.0 (Pollard et al. 2018, 2019), which contains ICU stays from 208 US hospitals in the Philips eICU telehealth programme in 2014–2015. The same rules were applied in the same order: age ≥ 18 years (ages recorded as "> 89" were coded as 90), first ICU stay per patient and then ICU length of stay ≥ 48 hours. Because eICU-CRD provides no absolute dates, the first stay was identified by ordering a patient's stays by hospital discharge year, health-system stay identifier and unit visit number, the ordering recommended by the database documentation as the closest available proxy for chronological order; stays of the same patient in different hospitals are linked by the database's unique patient identifier; stays with unknown hospital discharge status (569) or sex (9) were excluded. The external cohort comprised 54,696 stays from 205 hospitals (138,868 first adult stays; 55,274 with stays of at least 48 hours), with an in-hospital mortality of 11.9%. No eICU-CRD data were used for training, normalisation, early stopping, model selection or recalibration.

The landmark (prediction time) was 48 hours after ICU admission, and only observations timestamped between ICU admission and the landmark were used. The outcome was death before hospital discharge (MIMIC-IV hospital_expire_flag; eICU-CRD hospital discharge status "Expired"); because every included patient was alive and in the ICU at the landmark, every outcome event occurred after it. The target population is therefore patients still in the ICU 48 hours after admission to their first ICU stay, and the estimand is subsequent in-hospital death in that population. Patients who died, were discharged from the ICU or were transferred out of the ICU before 48 hours are excluded by design, including those who remained in hospital; the model is not intended for them. ICU length of stay was computed from the recorded ICU admission and discharge times (MIMIC-IV icustays; eICU-CRD unit discharge offset), so an ICU stay ending exactly at 48 hours was included. In MIMIC-IV, of 65,366 first ICU stays of adult patients, 31,142 lasted at least 48 hours and form the cohort; of the 34,224 excluded, 1,994 ended with death in the ICU before 48 hours, 926 ended with the patient leaving the ICU before 48 hours and dying later in hospital, 31,292 ended with the patient leaving the ICU before 48 hours and surviving to hospital discharge, and 12 had missing length of stay or discharge status.

## 3.2 Ethics

Both databases are de-identified and released under the PhysioNet Credentialed Health Data License v1.5.0; access was obtained through PhysioNet credentialing, which requires human-subjects research training and signed data use agreements. The creation of MIMIC-IV was reviewed by the Institutional Review Board at Beth Israel Deaconess Medical Center, which waived informed consent. The use of eICU-CRD is exempt from institutional review board approval because of its retrospective design, the lack of direct patient intervention and its security schema, for which the re-identification risk was certified as meeting the Health Insurance Portability and Accountability Act Safe Harbor standard by an independent privacy expert (Pollard et al. 2018). Patient-level data remained within the credentialed environment (PhysioNet-hosted BigQuery, the authors' workstation and a rented GPU instance). There was no patient or public involvement in this study. The study was not preregistered, and no protocol was published.

## 3.3 Leakage audits

Input timing. MIMIC-IV diagnosis codes (diagnoses_icd) carry no timestamp and are assigned at the hospital-admission level, typically at discharge, so they cannot be verified as available at 48 hours and were excluded. Vital signs, laboratory results and medication administrations carry individual timestamps and were restricted to the observation window.

Node features. A second audit found that the visit node of the previous version contained, besides age and sex, the total ICU length of stay, known only at discharge. Because every stay lasts at least 48 hours and longer stays are associated with death, this feature leaked post-prediction information. In this study the visit node contains only age at admission and sex, z-scored with training statistics, and the auxiliary length-of-stay training target of the previous version was removed. All results supersede those of the previous version and its preprint (Uddin et al. 2026), which will be updated to this version. Table 2 lists every change.

## 3.4 Temporal validation design

MIMIC-IV shifts each patient's dates by a random patient-specific offset, but releases for each patient the three-year group (anchor_year_group: 2008–2010, 2011–2013, 2014–2016, 2017–2019, 2020–2022) into which the patient's anchor year falls. We assigned every stay the anchor-year group of its patient; because each patient contributes one stay, no admission is split across groups. The groups locate a stay only approximately in calendar time (an admission can fall in a calendar year adjacent to its group), so we describe results by anchor-year group rather than by calendar year. Stays in the 2008–2010, 2011–2013 and 2014–2016 groups formed the development period and were split 70/15/15, stratified by outcome with a fixed seed, into training (14,172), validation (3,037) and internal test (3,038) sets. Stays in the 2017–2019 (6,142) and 2020–2022 (4,753) groups were held out as temporal test sets. Normalisation statistics, early stopping, model selection and recalibration used development-period data only. Each patient contributes one stay, so all sets are patient-disjoint.

## 3.5 Graph representation (CT-HEG)

Each stay is a typed graph with one visit node, one node per vital-sign variable and one node per laboratory analyte observed in the window: 15 vital-sign variables (arterial and non-invasive systolic, diastolic and mean blood pressure, heart rate, respiratory rate, peripheral oxygen saturation ($SpO_2$), temperature, fraction of inspired oxygen ($FiO_2$), the three Glasgow Coma Scale components and fingerstick glucose) and 27 laboratory analytes (Online Resource 1; Fig. 1). Each observation is a directed edge from the visit node to its variable node; reverse edges carrying the same attribute are added at load time, giving two relations in two directions. Each edge carries the elapsed time divided by 48 and the observed value. Each variable node carries six summary statistics of its observations (last, mean, population standard deviation, minimum, maximum and count). A variable with no observation in the window contributes no node and no edge; no value is imputed, and the model sees only the variables actually measured. Medication nodes, used in the previous version, were excluded: they added no detectable discrimination in its ablations, and medication records are not harmonisable between the two databases.

Variable-aware representation (v3). The revision makes two changes together. First, each variable node stores the index of its variable in a fixed vocabulary, used to look up a learned variable embedding (Sect. 3.6). Second, normalisation is per variable: the location statistics (last, mean, minimum, maximum) are transformed by the signed logarithm $\mathrm{sign}(x)\cdot\log(1 + |x|)$, which is defined for zero and negative values, and the standard deviation and count by $\log(1 + x)$; each transformed statistic is then z-scored with the mean and standard deviation of that variable over the training graphs. Edge values are transformed by the signed logarithm and z-scored with the training statistics of the destination variable, and elapsed time is divided by 48. Normalised values are clipped to $[-10, 10]$ after z-scoring. Because the two changes were introduced together, their separate contributions are not identified (Sect. 5.1). In the previous architecture (v2), summary statistics were z-scored across all variables of a node type pooled, edge values per relation, and no variable identity was represented. MIMIC-IV and eICU-CRD graphs were built by identical code.

## 3.6 Model (CHIRP-Net)

Each node's summary statistics are mapped to the hidden dimension by a type-specific linear encoder with layer normalisation, and a learned node-type embedding is added; in v3 a learned variable embedding (one 80-dimensional vector per variable) is also added to every vital-sign and laboratory node. Four message-passing layers follow. In the heterogeneous model each layer applies a separate graph attention network v2 (GATv2) convolution (Brody et al. 2022) with four heads to each of the four edge types, with attention depending on the source node, destination node and two-dimensional edge attribute; messages are summed across relations, passed through dropout (0.25), added residually and normalised with a node-type-specific layer norm. The relation-collapsed (homogeneous, or shared) model keeps the type-specific encoders and embeddings but uses one shared convolution and one shared layer norm per layer, so attention at the visit node is normalised jointly across all observations. The mortality probability is read from the final visit-node representation by a two-layer perceptron. The hidden dimension was fixed at 80, the value selected on validation in the previous capacity grid, giving 223,121 parameters for the heterogeneous and 62,481 for the relation-collapsed v3 model.

### 3.7 Training protocol

All CHIRP-Net variants were trained under one protocol: AdamW (learning rate $3\times10^{-4}$, weight decay $3\times10^{-4}$), 500-step linear warm-up and per-step cosine decay over 80 epochs, at most 80 epochs with early stopping after 15 epochs without improvement in the validation area under the receiver operating characteristic curve (AUROC), restoration of the best-validation checkpoint, class-balanced sampling, batch size 32 and gradient-norm clipping at 1.0. Class-balanced sampling drew training stays with replacement with probability inversely proportional to class frequency, so that training batches contained deaths and survivors in roughly equal proportion; the binary cross-entropy loss itself was unweighted. Validation and test predictions were made on the original, unbalanced sets, so raw predicted probabilities overestimate risk; this is corrected only by the recalibration described in Sect. 3.10. Seeds 42–46 varied initialisation and sampling; the data split was fixed. The v2 architecture was trained in heterogeneous and relation-collapsed forms at two capacities and as a vital/laboratory-only variant (hidden dimension 80), the last of which has the same inputs as v3.

### 3.8 Baselines

All baselines used the same stays, splits and window. Tabular models used, for every variable, the last, mean, standard deviation, minimum, maximum and count over the window, plus age and sex. Logistic regression used train-median imputation with missingness indicators, winsorisation at the training 0.5th and 99.5th percentiles, standardisation and L2 regularisation; gradient boosting used scikit-learn's histogram gradient boosting (Pedregosa et al. 2011) with native missing-value handling (five seeds). The tabular models reported in the tables were fitted with fixed hyperparameters (logistic regression C = 0.1; gradient boosting learning rate 0.05, 31 leaves, L2 penalty 1.0, internal early stopping) on the harmonised variable set, so that identical models could be applied to eICU-CRD; prespecified versions with hyperparameters chosen on validation AUROC gave equivalent MIMIC-IV results (gradient boosting AUROC 0.907 vs 0.909 in 2017–2019 and 0.902 vs 0.903 in 2020–2022). GRU-D (hidden dimension 256, five seeds) received hourly binned vital-sign and laboratory values with masks and time since last observation, z-scored with training statistics, with age and sex appended to the final hidden state. The day-1 Sequential Organ Failure Assessment (SOFA; Vincent et al. 1996), Simplified Acute Physiology Score II (SAPS-II; Le Gall et al. 1993) and Oxford Acute Severity of Illness Score (OASIS; Johnson et al. 2013) were taken from the MIMIC-IV derived concept tables (Johnson et al. 2018; tables first_day_sofa, sapsii and oasis, computed from the first 24 hours of the ICU stay; all three were available for every stay) and mapped to probabilities by a one-variable logistic regression on the training stays. In eICU-CRD, the APACHE IVa predicted hospital mortality (Zimmerman et al. 2006) recorded in the database served as a clinical benchmark; it uses first-24-hour physiology, admission diagnosis, admission source and comorbidities, information not available to the other models, and was available for 48,184 stays (88.1%). The severity scores and APACHE IVa therefore use a 24-hour window and the learned models a 48-hour window; this asymmetry is intentional, since the scores are included as established clinical reference points rather than as like-for-like competitors.

## 3.9 Harmonisation of eICU-CRD

eICU-CRD variables were mapped to the MIMIC-IV definitions (Online Resource 1). Periodic vital signs in eICU-CRD are five-minute aggregates, whereas MIMIC-IV vital signs are charted about hourly; to keep observation counts comparable, the first value per variable source in each hour since ICU admission (hour k covering minutes 60k to 60k + 59 of the stay, with minute 2,880 assigned to hour 47) was retained. Hours are defined relative to ICU admission, not by wall-clock time, so time zones and daylight-saving transitions do not affect them; identical repeated laboratory records were de-duplicated. Arterial and non-invasive blood pressure were kept separate, as in MIMIC-IV. Temperature was taken in degrees Celsius only, because nurse charting stores each reading in both Celsius and Fahrenheit; $FiO_2$ recorded as a fraction was converted to a percentage; bedside glucose was mapped to fingerstick glucose; and troponin T to troponin. C-reactive protein was reported in mg/dL by some hospitals and in mg/L by others under the same nominal unit and was converted per result using its reported unit. Vital signs outside physiological plausibility limits were removed (Online Resource 1). The eICU-CRD limits were set during harmonisation and are wider than those of the MIMIC-IV pipeline for heart rate, blood pressure, respiratory rate and $SpO_2$ (for example, heart rate 0–300 versus 10–300 per minute), and fingerstick glucose was bounded in eICU-CRD (10–1,500 mg/dL) but not in MIMIC-IV; a small number of implausible eICU-CRD values may therefore remain, although the harmonised distributions were close (Online Resource 2). Laboratory values were not bounded in either database. A harmonisation report (Online Resource 2) compares coverage, observations per stay and the 5th, 50th and 95th percentiles of every variable; after harmonisation, medians and observation counts were close for most variables, while some analytes were measured in fewer eICU-CRD stays (for example Glasgow Coma Scale 65% vs 99%, international normalised ratio (INR) 50% vs 92%, lactate 36% vs 68%, troponin 5% vs 24%).

## 3.10 Statistical analysis

Prediction summaries. Each learned model was trained with five seeds. Unless stated otherwise, the primary estimand is the expected performance of a single trained model, summarised as the mean of the five per-seed metrics with their population standard deviation (SD); this describes variation over training runs. We also report a five-model ensemble, the average of the five seeds' predicted probabilities, which is the deployable form; ensemble results are labelled as such. Discrimination was measured by the AUROC and the area under the precision–recall curve (AUPRC); AUPRC is interpreted against each set's event rate, given in Table 1. Uncertainty. Patient-level 95% percentile bootstrap intervals (2,000 resamples) were computed for the seed-mean metrics. Paired comparisons used a patient-level bootstrap (2,000 resamples) in which, in each resample, the metric difference between two models was computed for each seed on the same resampled stays and averaged across seeds; the 2.5th and 97.5th percentiles gave the 95% confidence interval (CI), and the two-sided p-value was twice the smaller of the proportions of resampled differences $\leq 0$ and $\geq 0$ (capped at 1). Bootstrap random seeds were fixed (2026 and 12345). These intervals are conditional on the trained models and reflect test-set sampling only; training variability is described by the seed SD. Patient-level intervals do not account for clustering of patients within hospitals and therefore understate uncertainty about performance in a new hospital; hospital-level variation is reported separately (Sect. 4.5). Intervals were not adjusted for multiple comparisons, and subgroup, hospital and ICU-type analyses are exploratory. For the comparison of relation-specific with relation-collapsed message passing we used an equivalence margin of ±0.01 AUROC, chosen after the analysis as a difference unlikely to matter for discrimination; two models were called equivalent when the 95% paired CI lay entirely inside the margin (a conservative two one-sided test at the 2.5% level). Calibration. Each seed's predictions were recalibrated by Platt scaling (Platt 1999), a logistic regression of the outcome on logit(p) fitted on that seed's MIMIC-IV validation predictions, and the five recalibrated predictions were then averaged for ensemble-based analyses. Calibration was summarised by the Brier score, calibration-in-the-large (the intercept a of logit $P(y = 1) = a + \text{logit}(p)$, with logit(p) as an offset; $a > 0$ indicates under-prediction and $a < 0$ over-prediction), the calibration slope (the coefficient of a logistic regression of the outcome on logit(p)) and the expected calibration error over 10 equal-frequency bins, each computed per seed and reported as mean and SD. Statistical utility. Decision curves (Vickers and Elkin 2006) over threshold probabilities 0.05–0.50 and sensitivity, specificity and predictive values at alert rates of 5%, 10% and 20% were computed from the recalibrated ensemble prediction; they are exploratory, because no specific clinical action or threshold was defined.

Post hoc analyses. The v3 architecture was introduced after the v2 results showed that v2 underperformed the tabular baselines, and three diagnostic analyses of the eICU-CRD results (by hospital, by ICU type and by data availability) were added after the first external results. These analyses were not prespecified; we report v2 and v3 side by side under identical protocols and treat their comparison as exploratory.

## 3.11 Implementation

Models were implemented in Python 3.12.15 with PyTorch 2.14.1, PyTorch Geometric 2.8.1 (Fey and Lenssen 2019) and scikit-learn 1.9.1, and trained on one rented NVIDIA RTX 5090 GPU. Cohorts for eICU-CRD and the clinical scores were extracted with PhysioNet-hosted BigQuery. Code is available at https://github.com/nasiruddinstudents-ctrl/chirp-net-mimic-iv (release v2.0.0, commit 69793b9; https://github.com/nasiruddinstudents-ctrl/chirp-net-mimic-iv/releases/tag/v2.0.0; archived at https://doi.org/10.5281/zenodo.23969044), with pinned package versions, the SQL queries and the scripts for every table and figure. The study is reported according to the TRIPOD+AI statement (Transparent Reporting of a multivariable prediction model for Individual Prognosis Or Diagnosis, updated for artificial intelligence; Collins et al. 2024).

## 3.12 Use of large language models

Claude (Anthropic) was used to support drafting of the manuscript and development of analysis code, including the eICU-CRD extraction, harmonisation and v3 scripts; Grammarly was used for language editing. The authors reviewed, ran and verified all code and outputs, checked them against independent results (for example, cohort sizes and the prespecified baselines), and take full responsibility for the content.

## 4 Results

### 4.1 Cohorts

Development-period and temporal test sets were similar in age and sex (Table 1). The 2020–2022 anchor-year group had longer ICU stays (median 103 vs 89 hours in training) and higher mortality (15.1% vs 13.1%); this group overlaps the COVID-19 pandemic, but COVID-19 status was not measured. eICU-CRD patients were of similar age (median 66 years) and sex (54.5% male), with lower mortality (11.9%). Characteristics of the full MIMIC-IV cohort by outcome, including race and ethnicity, insurance and first care unit, are given in Online Resource 3.

### 4.2 Discrimination across periods and hospitals

The revised model (CHIRP-Net v3, heterogeneous) reached a mean single-model AUROC of 0.878 ± 0.004 on the internal test set, 0.900 ± 0.002 in the 2017–2019 group and 0.896 ± 0.002 in the 2020–2022 group (Table 3); point estimates did not decrease in later groups, although no stability margin was prespecified. Its AUROC exceeded those of the clinical severity scores by 0.15–0.19 in the temporal test periods (2020–2022: SOFA +0.159, SAPS-II +0.147, OASIS +0.169; all $p < 0.001$) and by 0.16–0.23 on the internal test set. Its AUROC exceeded that of logistic regression in both temporal periods (+0.012, 95% CI 0.005–0.019, and +0.010, 0.003–0.017) but not significantly on the internal set (+0.009, −0.001 to 0.020), and was below gradient boosting on all three MIMIC-IV test sets (internal −0.011, −0.018 to −0.005; 2017–2019 −0.008, −0.013 to −0.004; 2020–2022 −0.007, −0.011 to −0.003). AUPRC was 0.566, 0.610 and 0.616 on the three sets (event rates 13.1%, 12.7% and 15.1%), against 0.594, 0.641 and 0.641 for gradient boosting. GRU-D reached AUROC 0.870 on the internal test set and 0.894 and 0.892 in the two temporal test periods.

In pooled external validation in eICU-CRD, without retraining, v3 reached a mean single-model AUROC of 0.809 ± 0.009 (bootstrap 95% CI of the seed mean 0.804–0.813), 0.010 below gradient boosting (−0.014 to −0.006) and 0.080 above logistic regression (0.074–0.086); AUPRC was 0.418 at an event rate of 11.9% (gradient boosting 0.440; logistic regression 0.298). Discrimination varied more across seeds in eICU-CRD than in MIMIC-IV (SD 0.009 vs 0.002), and the five-model ensemble discriminated better than the average single model. Restricted to the 48,184 stays with an APACHE IVa prediction, so that all comparators share one denominator, the v3 ensemble reached AUROC 0.825 and AUPRC 0.454, versus 0.788 and 0.365 for APACHE IVa (differences 0.037, 95% CI 0.031–0.043, and 0.089, 0.076–0.102; both $p < 0.001$), and the gradient-boosting ensemble reached AUROC 0.825 and AUPRC 0.461. The 0.809 (single-model mean, all 54,696 stays) and 0.825 (ensemble, complete cases) estimates therefore differ in both prediction summary and denominator.

### 4.3 Variable identity and relation typing

The variable-aware representation, which combines a variable embedding with per-variable normalisation, was the largest change (Table 4; Fig. 2). With the same inputs, protocol and hidden dimension, the v2 vital/laboratory model reached AUROC 0.820, 0.841, 0.846 and 0.699 on the internal, 2017–2019, 2020–2022 and eICU-CRD sets, and v3 reached 0.878, 0.900, 0.896 and 0.809: gains of about 0.05–0.06 within MIMIC-IV and 0.11 externally.

With the pooled v2 representation, relation-specific message passing scored higher than relation-collapsed message passing at approximately matched capacity (v2: 0.813 vs 0.804 internal; 0.842 vs 0.835 and 0.832 vs 0.827 in the temporal periods), differences of 0.005–0.009 in the same direction as in the previous version. With the variable-aware representation, v3 heterogeneous and relation-collapsed models reached 0.878 vs 0.879, 0.900 vs 0.900 and 0.896 vs 0.897. Paired differences (heterogeneous minus relation-collapsed) were −0.001 (95% CI −0.005 to 0.003) internally, −0.000 (−0.003 to 0.002) in 2017–2019, −0.000 (−0.003 to 0.002) in 2020–2022 and +0.003 (0.001 to 0.005) in eICU-CRD. All four intervals lie inside the ±0.01 equivalence margin, so the two architectures were equivalent at that margin in every test set; the external difference was statistically detectable in 54,696 stays but small (0.003). The relation-collapsed v3 model has 62,481 parameters, about a quarter of the heterogeneous model's 223,121, so the comparison is not capacity-matched; it shows that the larger relation-specific model offered no meaningful advantage, not that relation typing can never help.

## 4.4 Calibration and exploratory decision analysis

Raw predicted probabilities overestimated risk, as expected from class-balanced training (eICU-CRD Brier score 0.174 for v3, versus 0.105 for a constant prediction at the observed mortality). Platt recalibration fitted on the MIMIC-IV validation set reduced the eICU-CRD Brier score to 0.088 for v3 and 0.089 for the relation-collapsed model, lower than the 0.097 of APACHE IVa, which is calibrated for US ICUs by design (APACHE IVa evaluated on its 48,184 complete cases, where a constant prediction has a Brier score of 0.106). Within MIMIC-IV, recalibration brought calibration-in-the-large to within ±0.05 of zero and the calibration slope to 0.90–0.98 on all three test sets, with expected calibration errors of 0.010–0.015 (from 0.17–0.19 raw) and Brier scores of 0.081, 0.073 and 0.085, close to those of gradient boosting (0.077, 0.070 and 0.082). In eICU-CRD, calibration-in-the-large after recalibration was −0.08 and the calibration slope 0.72, indicating predictions that were too extreme in the external population, that is, meaningful miscalibration after transport; the corresponding values for gradient boosting were −0.40 and 0.84. Local recalibration would be required before any use in a new setting. In exploratory decision-curve analysis (Fig. 4), recalibrated v3 and gradient boosting had nearly identical net benefit across threshold probabilities of 0.05–0.50, both well above SAPS-II, treat-all and treat-none; in eICU-CRD, v3 had slightly higher net benefit than gradient boosting at thresholds above 0.15. Flagging the 10% of stays at highest predicted risk identified 44–50% of deaths in the MIMIC-IV test sets with a positive predictive value of 0.62–0.67, and 40% of deaths with a positive predictive value of 0.48 in eICU-CRD, similar to gradient boosting (Online Resource 5). These ranking summaries do not show that an alert would change care or outcomes.

## 4.5 Diagnostic analyses of transportability

For the v2 model, the eICU-CRD AUROC was lower than that of gradient boosting in every ICU type (0.68–0.78 vs 0.79–0.86) and across hospitals (median per-hospital AUROC 0.730 vs 0.817 in 115 hospitals with at least 100 stays and 10 deaths), whereas gradient boosting on the same harmonised variables transferred well. The external performance loss of v2 was therefore not attributable to the harmonised data alone. With the variable-aware representation, the picture changed. Across the same 115 hospitals, the seed-averaged v3 prediction had a median per-hospital AUROC of 0.824 (interquartile range 0.776–0.858), versus 0.817 (0.772–0.853) for gradient boosting and 0.775 (0.745–0.807) for APACHE IVa, and its AUROC equalled or exceeded that of APACHE IVa in 74% of these hospitals (interquartile ranges describe the spread of hospital-level estimates, each based on few deaths). The remaining 90 hospitals had too few stays or deaths for a hospital-level estimate. By ICU type, v3 reached AUROC 0.80–0.87, higher than APACHE IVa in all eight unit types and higher than gradient boosting in seven of the eight (the exception being neurological ICUs, 0.810 vs 0.837; Fig. 3a). External discrimination also differed by data availability; these subgroups differ in case mix and measurement intensity as well as in missingness, so the comparisons are descriptive. In the 35,321 stays (65%) with a Glasgow Coma Scale recorded in the first 48 hours, v3 reached AUROC 0.852 (gradient boosting 0.856; APACHE IVa 0.788), but in the 19,375 stays without one it reached 0.775 (0.768; APACHE IVa 0.795), the only subgroup in which APACHE IVa, which records its own neurological assessment, had the higher AUROC. v3 also discriminated better in stays with an arterial line (0.840 vs 0.814 without) and with a complete core laboratory panel (0.826 vs 0.788), and reached 0.863 in the 11,189 stays with all three (Fig. 3b; Online Resource 4).

### 4.6 Effect of the length-of-stay leak

On the random split of the previous version, removing the leaked length-of-stay feature lowered the AUROC of the heterogeneous model without auxiliary loss from 0.853 to 0.834 ± 0.004 (five seeds) and that of the relation-collapsed model from 0.840 to 0.820 ± 0.008 (four seeds; the fifth run was not completed), a difference of about 0.02 for both architectures. This estimate is confounded by a concurrent change (the corrected runs also z-scored the visit features, which the previous version left raw), so it bounds rather than isolates the effect of the leak; the fifth relation-collapsed run was not completed because the random-split runs were stopped to free compute for the v3 experiments. The typing difference without the variable-aware representation (+0.015 on this split) persisted after the leak was removed.

## 5 Discussion

The main finding is that, for this event-graph model and task, the node representation mattered more than the message-passing architecture. Giving each vital-sign and laboratory node a variable embedding, together with per-variable normalisation, raised mean single-model AUROC by about 0.05 within MIMIC-IV and by 0.11 in pooled external validation. Because the two components were introduced together, the ablation does not show which of them is responsible. A plausible mechanism, not demonstrated here, is that without either a node is a vector of six summary statistics on a pooled scale, so the model must infer from value ranges alone whether it is looking at a heart rate or an oxygen saturation; that inference would be fragile when value distributions shift between hospitals, consistent with the external loss of v2 being much larger than its internal deficit.

The second finding concerns how ablations are read. With the pooled representation, relation-specific message passing helped, and the previous version, with careful capacity matching, concluded that relation typing improves discrimination. With the variable-aware representation, the two architectures were equivalent within ±0.01 AUROC in every test set, and a relation-collapsed model with a quarter of the parameters sufficed. One reading, consistent with but not proven by these results, is that separate convolutions for vital-sign and laboratory relations partly compensated for the missing identity by letting the model at least distinguish the two observation types. An architectural benefit measured against an impoverished representation can therefore be a substitute for that representation rather than an independent effect; ablations of this kind should be interpreted against the strongest available representation.

Third, the corrected model was competitive with strong baselines but did not surpass gradient boosting, which remained 0.007–0.011 AUROC ahead in paired tests, with nearly identical calibration and net benefit, consistent with the strength of tree ensembles on tabular summaries (Grinsztajn et al. 2022). The previous version's apparent advantage over its baselines rested partly on a logistic regression that pooled statistics across variables (AUROC 0.769); a properly constructed logistic regression reached 0.887–0.888 in the temporal test periods. Weak baselines are thus a third evaluation hazard alongside leakage and capacity confounding. The graph model's advantages over gradient boosting are of a different kind: it represents individual observations and their timing without hand-crafted aggregation, accommodates additional entity types without redesigning a feature table, and its seed-averaged prediction matched that of gradient boosting in eICU-CRD stays with an APACHE IVa prediction (AUROC 0.825 for both).

Fourth, discrimination held across anchor-year groups and in pooled external validation. Point estimates did not decrease from the development period to the 2017–2019 and 2020–2022 groups, despite longer stays and higher mortality in the latter, and in eICU-CRD the ensemble's AUROC exceeded that of APACHE IVa, which uses admission diagnosis and comorbidities that the model never sees. Calibration, however, did not transport fully: after development-only recalibration the external calibration slope was 0.72, and performance depended on data availability, so local recalibration and prospective silent evaluation would be needed before any clinical use. The recalibrated external Brier score (0.088, all stays) was lower than that of APACHE IVa (0.097, complete cases), but the two were computed on different denominators and do not establish better clinical performance. External discrimination varied across seeds, and averaging seeds improved it; ensembling is a cheap remedy that should be considered for deployment.

## 5.1 Limitations

The variable embedding and per-variable normalisation were introduced together, and a factorial ablation separating them was not performed, so their individual contributions are unknown; the heterogeneous and relation-collapsed v3 models were not capacity-matched. The v3 architecture was designed after v2's deficit against the tabular baselines was observed; although it was trained under the unchanged protocol with the previously selected hidden dimension and no further tuning, its development was not prespecified, and the eICU-CRD diagnostics were likewise post hoc. MIMIC-IV offers only anchor-year groups, so the temporal split is approximate and calendar-year claims are not possible. The target population excludes patients who left the ICU or died before 48 hours, and no sensitivity analysis was performed for patients alive in hospital but outside the ICU at 48 hours. eICU-CRD covers 2014–2015 and US hospitals only; its harmonisation required choices (hourly thinning, unit handling), and some analytes were measured much less often, which may underestimate or overestimate transportability to other settings. External discrimination depended on data availability, particularly on whether the Glasgow Coma Scale was charted, so performance in settings that record neurological assessment less systematically may be lower. External uncertainty intervals are patient-level and do not account for clustering within hospitals, and hospital-level estimates were available only for 115 of 205 hospitals. Calibration metrics for the learned models in eICU-CRD were computed on all stays rather than on the APACHE IVa complete cases. The comparison with APACHE IVa is limited to stays with an APACHE prediction, and APACHE IVa uses 24-hour data whereas the other models use 48 hours. Comparisons with GRU-D were descriptive, because they come from a separate evaluation run. Medication data were not used, and the auxiliary analyses of the previous version (edge-value and timestamp ablations, subgroup analyses) were not repeated for v3. Attention weights were not used as explanations, no causal claim is made, and the findings concern one model family, one landmark and one outcome.

## 6 Conclusions

In a leakage-audited, temporally and externally validated evaluation of an event-graph model for post-48-hour ICU mortality, a variable-aware node representation mattered more than relation-specific message passing, whose apparent benefit was no longer detectable once that representation was in place. The corrected model's discrimination held across MIMIC-IV anchor-year groups and exceeded that of APACHE IVa in pooled external validation, but it did not outperform gradient boosting, and its external calibration required correction. For models of this kind, input timing and node features should be audited for leakage, comparisons should include properly constructed tabular baselines, architectural ablations should be interpreted against the strongest available representation, and validation should span time and sites before gains are attributed to graph structure. The model is not ready for clinical use.

## Statements and Declarations

**Funding:** No funding was received for conducting this study.

**Competing interests:** The authors have no relevant financial or non-financial interests to disclose.

**Ethics approval:** Secondary analysis of de-identified data (MIMIC-IV v3.1; eICU-CRD v2.0) under the PhysioNet Credentialed Health Data License; see Methods 3.2.

**Consent to participate / for publication:** Not applicable.

**Data availability:** MIMIC-IV v3.1 (https://doi.org/10.13026/kpb9-mt58) and eICU-CRD v2.0 (https://doi.org/10.13026/C2WM1R) are available to credentialed researchers on PhysioNet. The data use agreements prohibit redistribution, so no patient-level data are distributed; the released code regenerates the cohorts, splits and normalisation statistics deterministically.

**Code availability:** https://github.com/nasiruddinstudents-ctrl/chirp-net-mimic-iv, release v2.0.0 (commit 69793b9), archived at https://doi.org/10.5281/zenodo.23969044; BSD 3-Clause License. Per-patient predictions, graphs and model checkpoints derived from MIMIC-IV and eICU-CRD are not released because the data use agreements prohibit redistribution.

**Author contributions:** Mohammad Nasir Uddin: Conceptualization, Methodology, Software, Formal analysis, Investigation, Data curation, Validation, Visualization, Project administration, Writing – original draft, Writing – review & editing. Eklachur Rahman Bhuiyan: Methodology, Investigation, Validation, Writing – original draft, Writing – review & editing. Rahnuma Tabassum Orpita: Methodology, Software, Investigation, Validation, Writing – original draft, Writing – review & editing. E.R.B. and R.T.O. contributed these roles to the earlier version of the study on which this work builds. All authors read and approved the final manuscript and agree to be accountable for all aspects of the work.

## Acknowledgements

The authors used AI writing assistance (Claude, Anthropic; Grammarly) for manuscript drafting, editing and structural revision. All experimental design, code, results, data analysis and scientific conclusions are the authors' own, and the authors take full responsibility for the accuracy and integrity of the work.

# Tables

### Table 1. Characteristics of the development, temporal test and external validation cohorts.

| Set | Stays | Age, median (IQR) | Male, % | ICU LOS, h, median | Deaths, n (%) |
|---|---|---|---|---|---|
| MIMIC-IV training (2008–2016) | 14,172 | 68 (56–79) | 55.7 | 89 | 1,860 (13.1) |
| MIMIC-IV validation (2008–2016) | 3,037 | 67 (56–79) | 55.9 | 92 | 399 (13.1) |
| MIMIC-IV internal test (2008–2016) | 3,038 | 67 (55–79) | 54.4 | 88 | 399 (13.1) |
| MIMIC-IV 2017–2019 | 6,142 | 67 (55–77) | 57.1 | 95 | 782 (12.7) |
| MIMIC-IV 2020–2022 | 4,753 | 66 (55–76) | 58.9 | 103 | 720 (15.1) |
| eICU-CRD (205 hospitals, 2014–2015) | 54,696 | 66 (54–77) | 54.5 | 87 | 6,512 (11.9) |

ICU length of stay includes time after the 48-hour prediction point and was not used as a model input. IQR, interquartile range; LOS, length of stay.

### Table 2. Changes from the previous version of this study.

| Item | Previous version | This study |
|---|---|---|
| Visit-node features | Age, sex, total ICU length of stay (leak) | Age at admission and sex |
| Auxiliary training target | Total length of stay (weight 0.05) | None |
| Variable representation | No variable identity; statistics z-scored across variables pooled | Learned variable embedding and per-variable normalisation |
| Medication relation | In the primary model | Excluded |
| Validation | One random split of MIMIC-IV | Temporal (three periods) and external (eICU-CRD, 205 hospitals) |
| Baselines | Untuned; logistic regression pooled statistics across variables; no per-patient predictions | Per-variable gradient boosting and logistic regression, GRU-D, SOFA, SAPS-II, OASIS, APACHE IVa; paired tests |
| Calibration | Raw probabilities only | Before and after recalibration; decision curves |

### Table 3. Discrimination (AUROC) across test sets.

| Model | Internal (n=3,038) | 2017–2019 (n=6,142) | 2020–2022 (n=4,753) | eICU-CRD (n=54,696) |
|---|---|---|---|---|
| CHIRP-Net v3, heterogeneous | 0.878 ± 0.004 [0.862–0.894] | 0.900 ± 0.002 [0.890–0.910] | 0.896 ± 0.002 [0.886–0.906] | 0.809 ± 0.009 [0.804–0.813] |
| CHIRP-Net v3, relation-collapsed | 0.879 ± 0.005 [0.862–0.894] | 0.900 ± 0.001 [0.890–0.910] | 0.897 ± 0.001 [0.886–0.907] | 0.805 ± 0.010 [0.800–0.810] |
| Gradient boosting | 0.890 ± 0.001 [0.874–0.904] | 0.909 ± 0.001 [0.899–0.918] | 0.903 ± 0.001 [0.894–0.913] | 0.819 ± 0.002 [0.814–0.824] |
| Logistic regression | 0.870 [0.852–0.886] | 0.888 [0.876–0.900] | 0.887 [0.874–0.898] | 0.729 [0.722–0.735] |
| GRU-D | 0.870 [0.854–0.886] | 0.894 [0.883–0.904] | 0.892 [0.881–0.903] | — |
| CHIRP-Net v2, vital/lab | 0.820 | 0.841 | 0.846 | 0.699 |

| Model | Internal (n=3,038) | 2017–2019 (n=6,142) | 2020–2022 (n=4,753) | eICU-CRD (n=54,696) |
|---|---|---|---|---|
| SAPS-II | 0.722 | 0.743 | 0.750 | — |
| SOFA (day 1) | 0.653 | 0.729 | 0.737 | — |
| OASIS | 0.699 | 0.712 | 0.727 | — |
| APACHE IVa (complete cases, n=48,184) | — | — | — | 0.788 |

Five-seed mean ± SD (population SD) and 95% bootstrap interval of the seed mean; single-fit models show the interval only. Tabular models use the harmonised variable set. GRU-D and the v2 model come from the prespecified v2 evaluation. AUPRC (CHIRP-Net v3 heterogeneous / gradient boosting): internal 0.566 / 0.594; 2017–2019 0.610 / 0.641; 2020–2022 0.616 / 0.641; eICU-CRD 0.418 / 0.440. Paired AUROC differences, v3 heterogeneous minus comparator, internal / 2017–2019 / 2020–2022 / eICU-CRD: gradient boosting −0.011 / −0.008 / −0.007 / −0.010 (all 95% CIs exclude zero); logistic regression +0.009 (not significant) / +0.012 / +0.010 / +0.080; relation-collapsed −0.001 / −0.000 / −0.000 / +0.003. AUPRC, area under the precision–recall curve; AUROC, area under the receiver operating characteristic curve; GRU-D, gated recurrent unit with decay; OASIS, Oxford Acute Severity of Illness Score; SAPS, Simplified Acute Physiology Score; SOFA, Sequential Organ Failure Assessment

### Table 4. Variable-aware representation and relation typing (test AUROC).

| Architecture | Internal | 2017–2019 | 2020–2022 | eICU-CRD |
|---|---|---|---|---|
| v2 (pooled representation), relation-specific | 0.813 | 0.842 | 0.832 | — |
| v2 (pooled representation), relation-collapsed | 0.804 | 0.835 | 0.827 | — |
| v2, vital/laboratory, relation-specific | 0.820 | 0.841 | 0.846 | 0.699 |
| v3 (variable-aware), relation-specific | 0.878 | 0.900 | 0.896 | 0.809 |
| v3 (variable-aware), relation-collapsed | 0.879 | 0.900 | 0.897 | 0.805 |

v2 heterogeneous and relation-collapsed rows are the hidden-80 and hidden-192 configurations with medication nodes and without the auxiliary loss (approximately matched capacity, as in the previous version); the v2 vital/lab row has the same inputs as v3. Five-seed means.

### Table 5. External validation in eICU-CRD: comparison with APACHE IVa and calibration.

| Model | AUROC | AUPRC | Brier, raw | Brier, recalibrated |
|---|---|---|---|---|
| A. Complete cases (n = 48,184), ensemble predictions | | | | |
| CHIRP-Net v3, five-model ensemble | 0.825 | 0.454 | n.c. | n.c. |
| Gradient boosting, five-model ensemble | 0.825 | 0.461 | n.c. | n.c. |
| APACHE IVa | 0.788 | 0.365 | 0.097 | — |
| B. All stays (n = 54,696), mean of single models | | | | |
| CHIRP-Net v3 | 0.809 | 0.418 | 0.174 | 0.088 |
| Gradient boosting | 0.819 | 0.440 | 0.086 | 0.088 |
| CHIRP-Net v2, vital/laboratory | 0.699 | 0.252 | 0.177 | 0.101 |

Complete cases: 48,184 stays with an APACHE IVa prediction. v3 minus APACHE IVa: AUROC +0.037 (95% CI 0.031–0.043), AUPRC +0.089 (0.076–0.102). Recalibration used the Platt map fitted on the MIMIC-IV validation set only. Brier score of a constant prediction at the observed mortality: 0.106 (panel A) and 0.105 (panel B). Brier scores in panels A and B are on different denominators and should not be compared directly. n.c., not computed on the complete-case subset.

# Figures

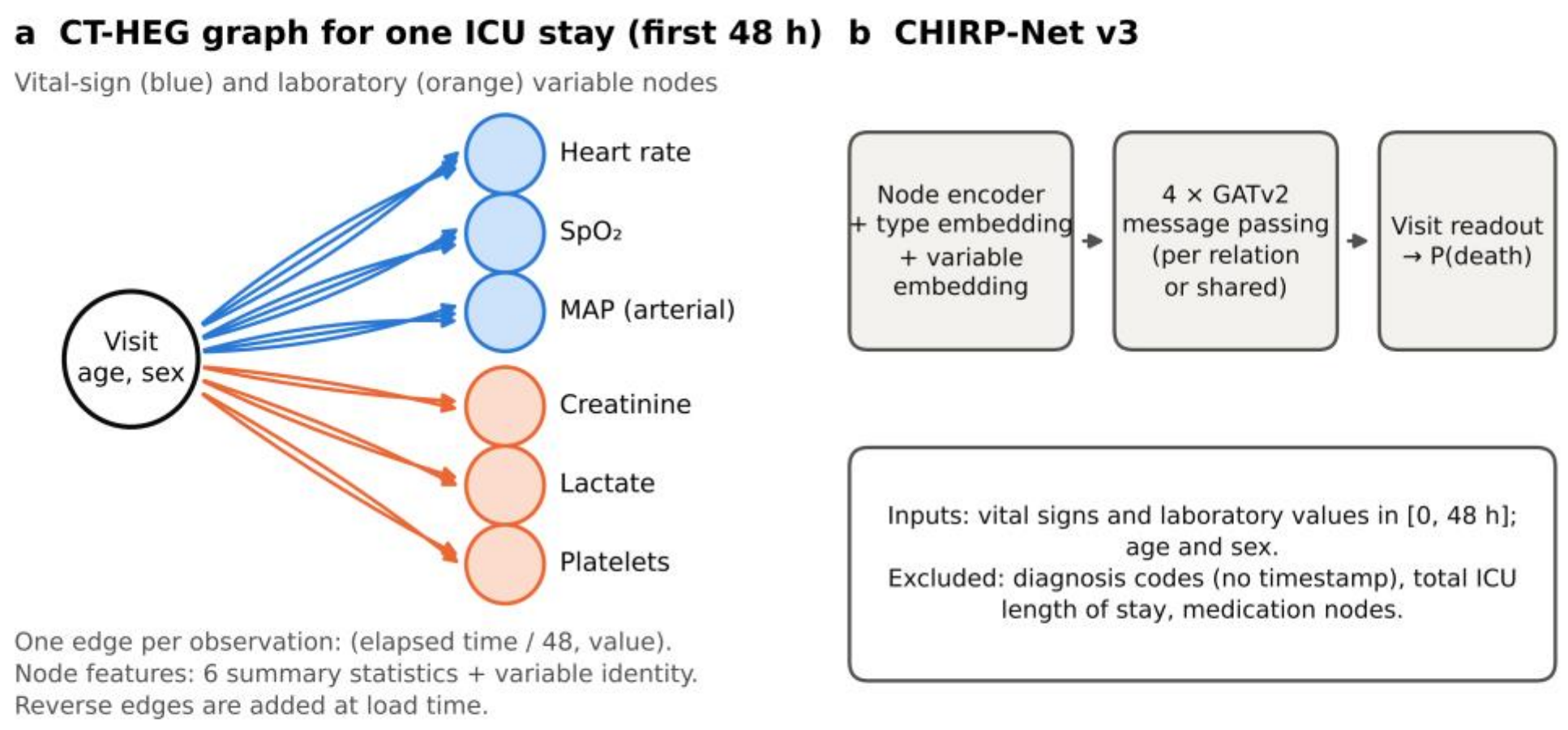

**Fig. 1** CT-HEG and CHIRP-Net v3. a A stay is a visit node (age and sex) linked to vital-sign and laboratory variable nodes by one directed edge per observation in the first 48 hours, each carrying normalised elapsed time and value; reverse edges are added at load time. b Each node is encoded from its six summary statistics plus a node-type embedding and, in v3, a per-variable embedding, followed by four GATv2 message-passing layers (relation-specific or shared) and a mortality readout from the visit node. GATv2, graph attention network v2; MAP, mean arterial pressure; $SpO_2$, peripheral oxygen saturation

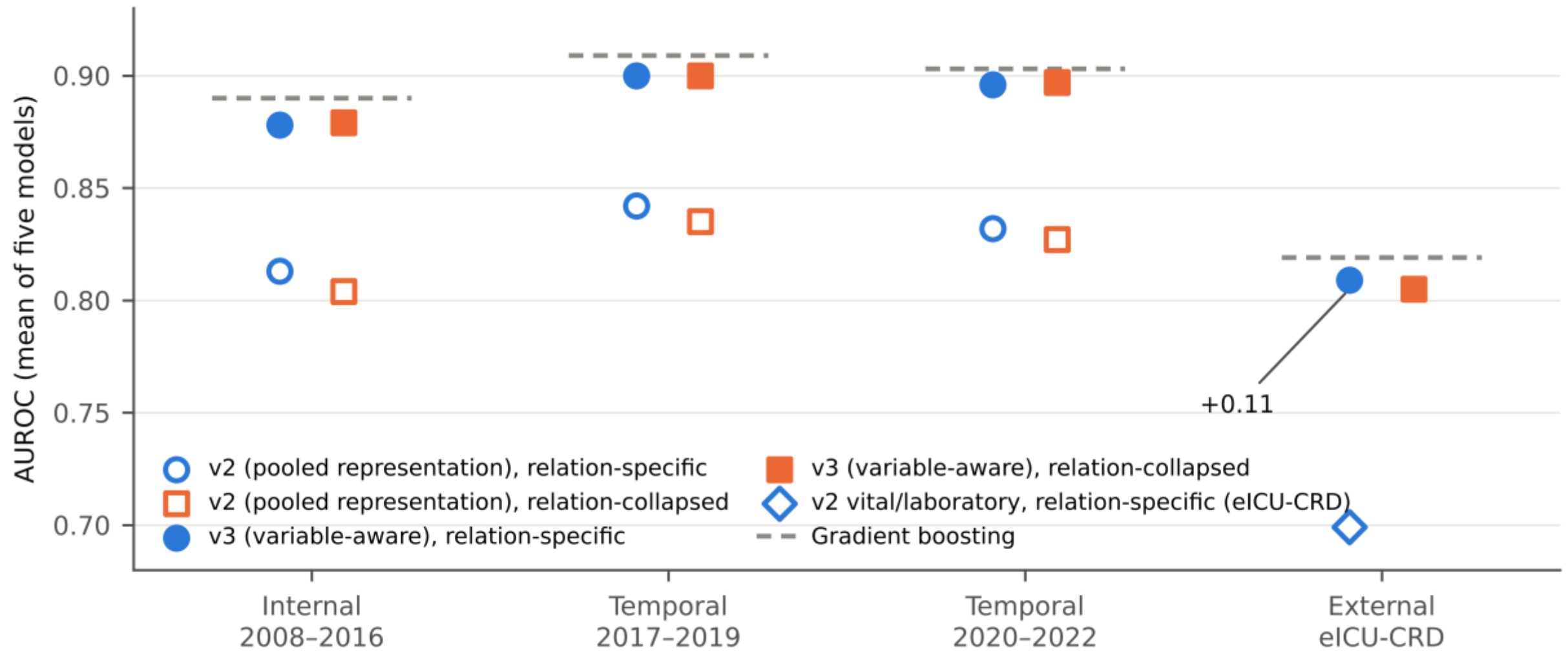

**Fig. 2** Variable-aware representation and relation typing. Test-set AUROC (mean of five single models) of relation-specific (circles) and relation-collapsed (squares) message passing with the pooled v2 representation (open markers) and the variable-aware v3 representation (filled markers); the open diamond is the v2 vital/laboratory model, which has the same inputs as v3, in eICU-CRD. Dashed lines mark gradient boosting. v2 was not evaluated in eICU-CRD

with medication nodes because medication records are not harmonisable. AUROC, area under the receiver operating characteristic curve

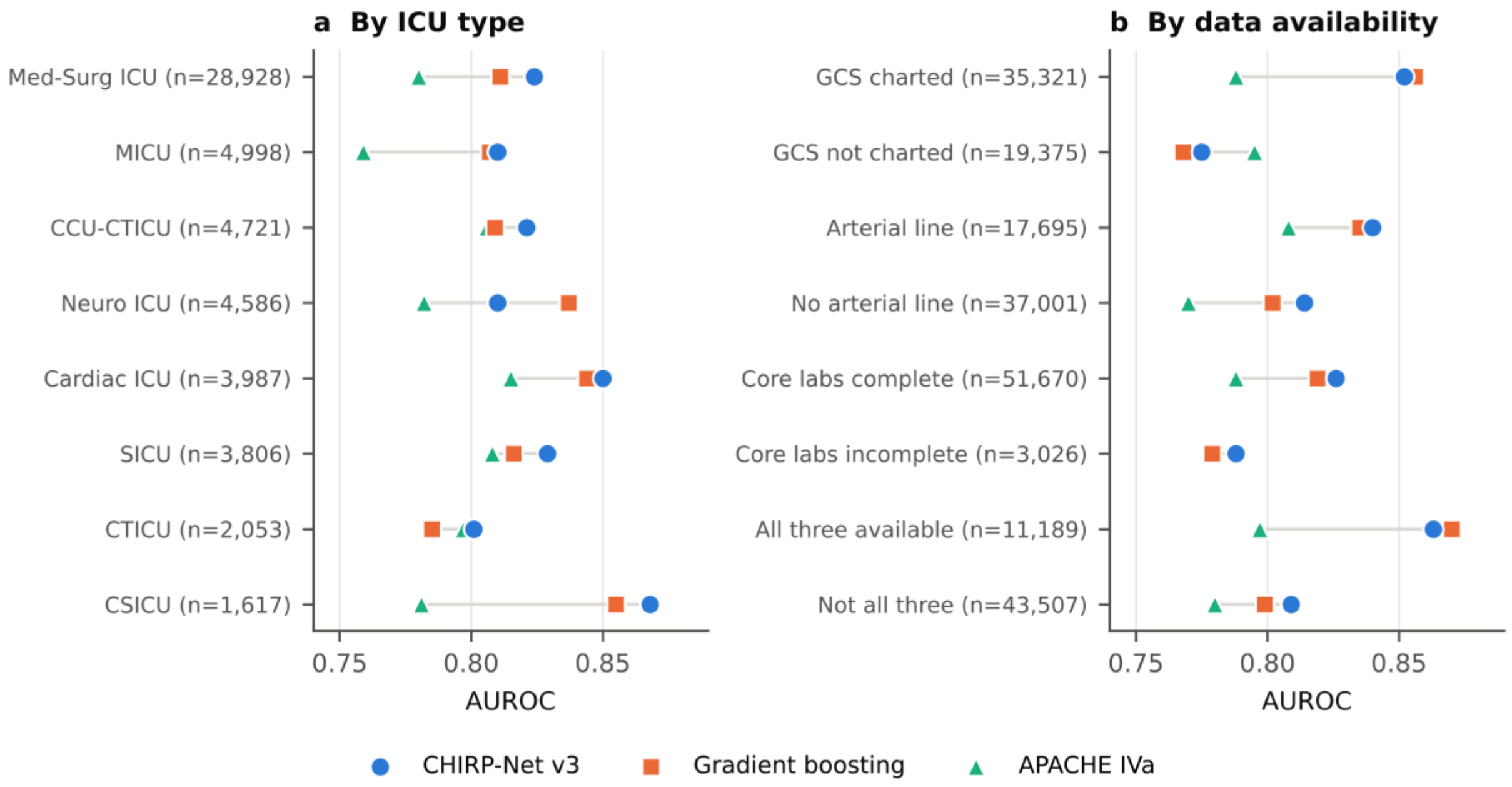

**Fig. 3** External discrimination in eICU-CRD by ICU type (a) and by data availability in the first 48 hours (b). Markers show AUROC of the five-model CHIRP-Net v3 ensemble (circles), gradient boosting (squares) and APACHE IVa (triangles; stays with an APACHE IVa prediction). Core laboratory panel: creatinine, urea nitrogen, sodium, potassium, white blood cell count, platelets and haemoglobin. APACHE, Acute Physiology and Chronic Health Evaluation; GCS, Glasgow Coma Scale

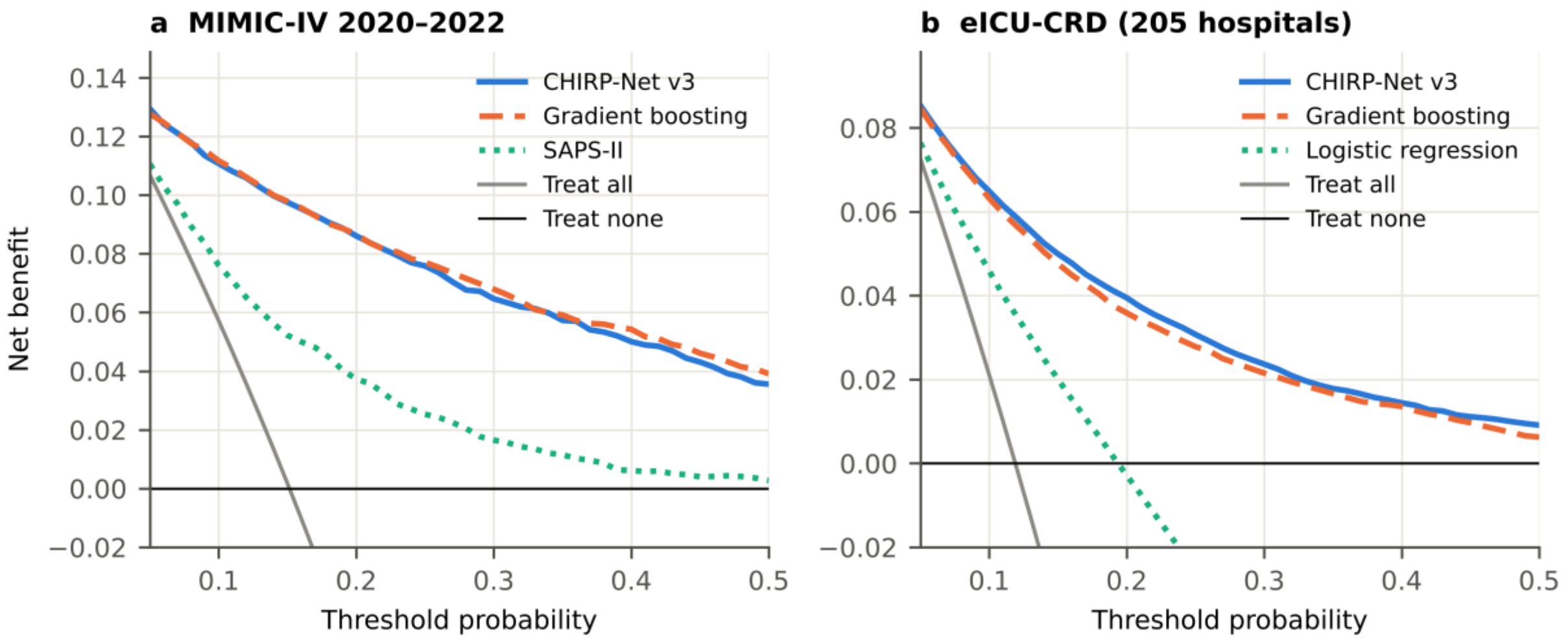

**Fig. 4** Decision curves for MIMIC-IV 2020–2022 (a) and eICU-CRD (b). Exploratory net benefit of the recalibrated five-model ensembles of CHIRP-Net v3 and gradient boosting, with SAPS-II (a) or logistic regression (b), against treating all and treating none, at threshold probabilities 0.05–0.50. SAPS, Simplified Acute Physiology Score

## References

Bignoumba N, Mellouli N, Ben Yahia S (2024) A new efficient alignment-driven neural network for mortality prediction from irregular multivariate time series data. Expert Syst Appl 238:122148. https://doi.org/10.1016/j.eswa.2023.122148

Brody S, Alon U, Yahav E (2022) How attentive are graph attention networks? In: International Conference on Learning Representations (ICLR). arXiv:2105.14491

Che Z, Purushotham S, Cho K, Sontag D, Liu Y (2018) Recurrent neural networks for multivariate time series with missing values. Sci Rep 8:6085. https://doi.org/10.1038/s41598-018-24271-9

Chen J, Yin C, Wang Y, Zhang P (2024) Predictive modeling with temporal graphical representation on electronic health records. In: Proceedings of the Thirty-Third International Joint Conference on Artificial Intelligence (IJCAI-24), pp 5763–5771. https://doi.org/10.24963/ijcai.2024/637

Cifci MA, Öney B, Yildirim F, Yilmaz Başer H, Zontul M (2025) Interpretable adaptive graph fusion network for mortality and complication prediction in ICUs. Diagnostics 15:2825. https://doi.org/10.3390/diagnostics15222825

Collins GS, Moons KGM, Dhiman P, Riley RD, Beam AL, Van Calster B et al (2024) TRIPOD+AI statement: updated guidance for reporting clinical prediction models that use regression or machine learning methods. BMJ 385:e078378. https://doi.org/10.1136/bmj-2023-078378

Fey M, Lenssen JE (2019) Fast graph representation learning with PyTorch Geometric. In: ICLR Workshop on Representation Learning on Graphs and Manifolds. arXiv:1903.02428

Grinsztajn L, Oyallon E, Varoquaux G (2022) Why do tree-based models still outperform deep learning on typical tabular data? In: Advances in Neural Information Processing Systems 35 (Datasets and Benchmarks Track). arXiv:2207.08815

Harutyunyan H, Khachatrian H, Kale DC, Ver Steeg G, Galstyan A (2019) Multitask learning and benchmarking with clinical time series data. Sci Data 6:96. https://doi.org/10.1038/s41597-019-0103-9

Johnson AEW, Kramer AA, Clifford GD (2013) A new severity of illness scale using a subset of Acute Physiology And Chronic Health Evaluation data elements shows comparable predictive accuracy. Crit Care Med 41:1711–1718. https://doi.org/10.1097/CCM.0b013e31828a24fe

Johnson AEW, Stone DJ, Celi LA, Pollard TJ (2018) The MIMIC Code Repository: enabling reproducibility in critical care research. J Am Med Inform Assoc 25:32–39. https://doi.org/10.1093/jamia/ocx084

Johnson AEW, Bulgarelli L, Shen L, Gayles A, Shammout A, Horng S et al (2023) MIMIC-IV, a freely accessible electronic health record dataset. Sci Data 10:1. https://doi.org/10.1038/s41597-022-01899-x

Johnson A, Bulgarelli L, Pollard T, Gow B, Moody B, Horng S et al (2024) MIMIC-IV (version 3.1). PhysioNet. https://doi.org/10.13026/kpb9-mt58

Kapoor S, Narayanan A (2023) Leakage and the reproducibility crisis in machine-learning-based science. Patterns 4:100804. https://doi.org/10.1016/j.patter.2023.100804

Kaufman S, Rosset S, Perlich C, Stitelman O (2012) Leakage in data mining: formulation, detection, and avoidance. ACM Trans Knowl Discov Data 6:15. https://doi.org/10.1145/2382577.2382579

Le Gall JR, Lemeshow S, Saulnier F (1993) A new Simplified Acute Physiology Score (SAPS II) based on a European/North American multicenter study. JAMA 270:2957–2963. https://doi.org/10.1001/jama.1993.03510240069035

Li S, Cheng H, Li R, Li W (2024) Time-aware heterogeneous graph transformer with adaptive attention merging for health event prediction. arXiv:2404.14815

Pedregosa F, Varoquaux G, Gramfort A, Michel V, Thirion B, Grisel O et al (2011) Scikit-learn: machine learning in Python. J Mach Learn Res 12:2825–2830

Platt J (1999) Probabilistic outputs for support vector machines and comparisons to regularized likelihood methods. In: Smola AJ, Bartlett P, Schölkopf B, Schuurmans D (eds) Advances in large margin classifiers. MIT Press, Cambridge, pp 61–74

Pollard TJ, Johnson AEW, Raffa JD, Celi LA, Mark RG, Badawi O (2018) The eICU Collaborative Research Database, a freely available multi-center database for critical care research. Sci Data 5:180178. https://doi.org/10.1038/sdata.2018.178

Pollard T, Johnson A, Raffa J, Celi LA, Badawi O, Mark R (2019) eICU Collaborative Research Database (version 2.0). PhysioNet. https://doi.org/10.13026/C2WM1R

Rahman MT, Al Olaimat M, Bozdag S (2026) DyGraphTrans: a temporal graph representation learning framework for modeling disease progression from electronic health records. bioRxiv. https://doi.org/10.64898/2026.01.28.702347

Rocheteau E, Tong C, Veličković P, Lane N, Liò P (2021) Predicting patient outcomes with graph representation learning. arXiv:2101.03940

Shi T, Xu H, Ma J, Kong G (2024) ICU-TGNN: a hybrid multitask transformer and graph neural network model for predicting clinical outcomes of patients in the ICU. In: 2024 IEEE International Conference on Systems, Man, and Cybernetics (SMC), pp 1761–1766. https://doi.org/10.1109/SMC54092.2024.10831750

Shukla SN, Marlin BM (2018) Modeling irregularly sampled clinical time series. arXiv:1812.00531

Shukla SN, Marlin BM (2021) Multi-time attention networks for irregularly sampled time series. In: International Conference on Learning Representations (ICLR)

Uddin MN et al (2026) CT-HEG: a bidirectional, timestamp-attributed event graph for ICU in-hospital mortality prediction – an architectural ablation study. arXiv:2608.02663

Vickers AJ, Elkin EB (2006) Decision curve analysis: a novel method for evaluating prediction models. Med Decis Making 26:565–574. https://doi.org/10.1177/0272989X06295361

Vincent JL, Moreno R, Takala J, Willatts S, De Mendonça A, Bruining H et al (1996) The SOFA (Sepsis-related Organ Failure Assessment) score to describe organ dysfunction/failure. Intensive Care Med 22:707–710. https://doi.org/10.1007/BF01709751

Wang Y, Jiang S, Wang C, Wan Z, Fan Y, Li K, Huang L, Zhou F (2026) MedGAITS: a graph autoencoder network for modeling irregular time series data in electronic medical records. Health Inf Sci Syst 14:40. https://doi.org/10.1007/s13755-026-00434-1

Wanyan T, Honarvar H, Azad A, Ding Y, Glicksberg BS (2021) Deep learning with heterogeneous graph embeddings for mortality prediction from electronic health records. Data Intell 3:329–339. https://doi.org/10.1162/dint_a_00097

Zimmerman JE, Kramer AA, McNair DS, Malila FM (2006) Acute Physiology and Chronic Health Evaluation (APACHE) IV: hospital mortality assessment for today's critically ill patients. Crit Care Med 34:1297–1310. https://doi.org/10.1097/01.CCM.0000215112.84523.F0